\pdfoutput=1
\PassOptionsToPackage{dvipsnames}{xcolor}
\documentclass[11pt]{article}

\usepackage[]{acl}

\usepackage{times}
\usepackage{latexsym}

\usepackage[T1]{fontenc}
\usepackage[utf8]{inputenc}
\usepackage{microtype}
\usepackage{inconsolata}
\usepackage{hyperref}
\usepackage{makecell}
\usepackage{subcaption}

\usepackage{tabularx}
\usepackage[dvipsnames]{xcolor}

\usepackage{pifont}
\usepackage{graphicx}
\usepackage{amsbsy}
\usepackage{vcell}
\usepackage{multirow}
\usepackage{amsmath}
\usepackage[ruled,vlined]{algorithm2e}
\SetInd{0.2em}{0.3em}
\SetArgSty{textup}
\usepackage{lipsum}

\usepackage{environ}
\usepackage{varwidth}

\usepackage{tcolorbox}
\tcbuselibrary{many,breakable}
\definecolor{carmine}{rgb}{0.59, 0.0, 0.09}
\definecolor{bubblegreen}{RGB}{103,184,104}
\definecolor{bubblegray}{RGB}{241,240,240}
\definecolor{tealblue}{rgb}{0.21, 0.46, 0.53}

\newtcolorbox{protip}[2][]{
fontupper = \linespread{.85}\fontsize{10pt}{12pt}\selectfont,
  colframe = gray, 
  colback  = white,
  coltitle = #2!20!black,  
  segmentation style={solid},
  boxrule=1pt,
  left=2pt,
  right=2pt,
  top=2pt,
  bottom = 2pt,
  middle = 2pt,
  arc=0pt,
  #1,
}

\newtcolorbox{context}[3][]
{fontupper=\linespread{.6}\selectfont,
after skip=3pt,
  colframe = gray,
  colback  = white,
  coltitle = #2!20!black,  
  segmentation style={solid},
  arc=2mm,
  boxrule=1pt,
  left=2pt,
  right=2pt,
  top=2pt,
  bottom = 2pt,
  middle = 2pt,
  #1,
}

\newtcolorbox{hs}[3][]
{fontupper=\linespread{.6}\selectfont,
after skip=3pt,
  colframe = carmine,
  colback  = white,
  coltitle = #2!20!black,  
  sharp corners = southwest,
  segmentation style={solid},
  arc=2mm,
  boxrule=1pt,
  left=2pt,
  right=2pt,
  top=2pt,
  bottom = 2pt,
  middle = 2pt,
  #1,
}

\newtcolorbox{cs}[3][]
{fontupper=\linespread{.6}\selectfont,
  colframe = teal,
  colback  = white,
  coltitle = #2!20!black,  
  sharp corners = southeast,
  segmentation style={solid},
  boxrule=1pt,
  arc=2mm,
  left=2pt,
  right=2pt,
  top=2pt,
  bottom = 2pt,
  middle = 2pt,
  #1,
}

\newtcolorbox{csreply}[3][]
{fontupper=\linespread{.6}\selectfont,
before skip=3pt,
  colframe = teal,
  colback  = white,
  coltitle = #2!20!black,  
  sharp corners = southeast,
  segmentation style={solid},
  boxrule=1pt,
  arc=2mm,
  left=2pt,
  right=2pt,
  top=2pt,
  bottom = 2pt,
  middle = 2pt,
  #1,
}

\newtcolorbox{inputmess}[3][]
{fontupper=\linespread{.6}\selectfont,
after skip=3pt,
  colframe = gray,
  colback  = white,
  coltitle = #2!20!black,  
  sharp corners = southwest,
  segmentation style={solid},
  boxrule=1pt,
  arc=2mm,
  left=2pt,
  right=2pt,
  top=2pt,
  bottom = 2pt,
  middle = 2pt,
  #1,
}

\newtcolorbox{reply}[3][]
{fontupper=\linespread{.6}\selectfont,
before skip=3pt,
  colframe = gray,
  colback  = white,
  coltitle = #2!20!black,  
  sharp corners = southeast,
  segmentation style={solid},
  boxrule=1pt,
  arc=2mm,
  left=2pt,
  right=2pt,
  top=2pt,
  bottom = 2pt,
  middle = 2pt,
  #1,
}

\newtcolorbox{replyx}[3][]
{fontupper=\linespread{.6}\selectfont,
before skip=3pt,
after skip = 3pt,
  colframe = gray,
  colback  = white,
  coltitle = #2!20!black,  
  sharp corners = southeast,
  segmentation style={solid},
  boxrule=1pt,
  arc=2mm,
  left=2pt,
  right=2pt,
  top=2pt,
  bottom = 2pt,
  middle = 2pt,
  #1,
}

\usepackage{placeins} 
\usepackage{booktabs}
\usepackage[normalem]{ulem}
\useunder{\uline}{\ul}{}
\usepackage{enumitem}

\title{NLP for Counterspeech against Hate:\\ A Survey and \emph{How-To} Guide \includegraphics[scale=0.045]{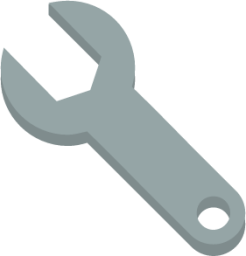} }

\author{
Helena Bonaldi\textsuperscript{1,2} \quad
Yi-Ling Chung\textsuperscript{3} \quad
Gavin Abercrombie\textsuperscript{4} \quad
Marco Guerini\textsuperscript{1} \\
\textsuperscript{1}Fondazione Bruno Kessler, Italy,
\textsuperscript{2} University of Trento, Italy, \\
\textsuperscript{3}The Alan Turing Institute,
\textsuperscript{4}The Interaction Lab, Heriot-Watt University \\
\texttt{hbonaldi@fbk.eu}, \texttt{ychung@turing.ac.uk},
\texttt{g.abercrombie@hw.ac.uk}, \texttt{guerini@fbk.eu} 
}

\date{}

\begin{document}

\maketitle
\begin{abstract}
In recent years, counterspeech has emerged as one of the most promising strategies to fight online hate. These non-escalatory responses tackle online abuse while preserving the freedom of speech of the users, and can have a tangible impact in reducing online and offline violence. Recently, there has been growing interest from the Natural Language Processing (NLP) community in addressing the challenges of analysing, collecting, classifying, and automatically generating counterspeech, to reduce the huge burden of manually producing it. In particular, researchers have taken different directions in addressing these challenges, thus providing a variety of related tasks and resources. In this paper, we provide a guide for doing research on counterspeech, by describing---with detailed examples---the steps to undertake, and providing best practices that can be learnt from the NLP studies on this topic. Finally, we discuss open challenges and future directions of counterspeech research in NLP.
\end{abstract}

\noindent{\color{red}Content warning: this paper contains unobfuscated examples some readers may find offensive}

\section{Introduction}

Online spaces provide fertile ground for the diffusion of hateful content, which is often interlinked with episodes of offline violence \cite{awan2016affinity}. 
Both witnessing and receiving hateful content can be detrimental to the mental health of victims and create a sense of insecurity \citep{saha_prevalence_2019, siegel2020online, dreissigacker2024online}, determining the need to mitigate hate.
In this context, counterspeech represents a promising strategy to oppose online hate, since it can be more effective than other moderation procedures \cite{benesch2014countering, schieb2016governing}, while also protecting free speech \cite{kiritchenko2020confronting}. Because of its potential effectiveness, counterspeech has been investigated by non-governmental organisations (NGOs) as a possible strategy to fight online hate. An example of hate speech (HS) and counterspeech (CS) from \citet{fanton2021human} is shown here:\footnote{Throughout the paper, examples are coded with the following colour boxes: {\color{carmine}red} for hate speech, {\color{teal}light blue} for counterspeech, and {\color{gray}grey} for everything else.}\looseness=-1

\begin{hs}{carmine}{}{}
    \small HS: Women are basically childlike, they remain this way most of their lives. Soft and emotional. It has devastated our once great patriarchal civilizations.
\end{hs}
\begin{csreply}{teal}{}{}
\small CS: Without softness and emotions there would be just brutality and cruelty. Not all women are soft and emotional and many men have these characteristics. To perpetuate these socially constructed gender profiles maintains patriarchal norms which oppress both men and women.
\end{csreply}

Given the amount of hateful content produced, an increasing number of Natural Language Processing (NLP) studies have begun to address the task of automatic counterspeech classification and generation. However, the settled definitions and best practices required to unify these efforts are still missing. While prior surveys largely focused on the effectiveness of deploying counterspeech in the real world ~\citep{chaudhary2021countering,adak2022mining,alsagheer2022counter,chung2023understanding}, we offer a complete step-by-step guide on how to conduct NLP research on counterspeech for both newcomers and experts. In particular, we extensively review existing NLP studies and resources on counterspeech, propose common concepts and best practices, and point out the limitations and open challenges of what has been done so far. 
After providing some background (\S\ref{sec:background}), the guide is articulated in three steps: task design, data selection and evaluation (\S\ref{sec:task_design}, \S\ref{sec:select_data}, \S\ref{sec:evalution}, respectively).\footnote{These sections contain practical recommendations
and best practices, marked with the spanner symbol \includegraphics[scale=0.03]{wrench.png}.} 
Finally, we discuss the open challenges in the field.
A complete description of the review methodology we used is provided in Appendix \ref{app:meth_review}.

\section{Background} \label{sec:background}
To better frame the concept of counterspeech we review definitions that have been proposed for it, the strategies that can be adopted, and several related tasks. 

\subsection{Definitions}

The most common definition of counterspeech is that of \citet{benesch2014countering} and \citet{schieb2016governing}, who identify it as non-aggressive textual feedback that uses credible evidence, factual arguments and alternative viewpoints. 
Other works have focused on the relational nature of counterspeech: it only exists in response to hate speech \cite{mathew2019thou, ashida2022towards}, challenging, condemning it or providing an alternative viewpoint \cite{vidgen2021introducing, hangartner2021empathy}. Also, it should
explicitly condemn hate or support an abused entity \cite{he2021racism, vidgen2020detecting}. Finally, it is consequences-oriented: it should discourage hate speech \cite{rieger2018hate} and aim to change what people think~\cite{qian-etal-2019-benchmark}.  

Although the terms \textit{counterspeech} and \textit{counter narrative} both rely on the idea that ``the strategic response to hate speech is more speech'' \cite{bielefeldt2011ohchr}, in the social sciences \textit{counter narratives} are representations that challenge dominant views in the areas of education, propaganda and public information \cite{benesch2016}.\footnote{We refer to \citet{chung2023understanding} for a more detailed analysis of this distinction.} 
Nevertheless, in the NLP studies included in this survey, these terms have been used interchangeably. 
Accordingly, we analyse works focusing on both ``counter narratives'' and ``counterspeech'', but use the latter term, which we consider to be more appropriate.\looseness=-1

\subsection{Strategy taxonomies} \label{sec:taxonomies}
Counterspeech can be distinguished by the strategy (or strategies) it employs. The most common taxonomy is that proposed by \citet{benesch2016}, who distinguish seven types of counterspeech: \textit{Presenting facts to correct misstatements or misperceptions}, \textit{Pointing out hypocrisy or contradictions}, \textit{Warning of consequences}, \textit{affiliation}, \textit{Denouncing hateful speech}, \textit{Humor and sarcasm} and \textit{Tone}. \citet{mathew2019thou} split the latter category into \textit{Positive} and \textit{Hostile language}, and \citet{conan-2019} add \textit{Counter-questions} on top of these. Other taxonomies have been proposed by \citet{qian-etal-2019-benchmark} and \citet{vidgen2020detecting}: see Appendix \ref{app:taxonomies_table} for more details.
However, not all strategies are equally effective: using \textit{Hostile tone} can backfire, or discourage other counterspeakers from joining a conversation \cite{benesch2016considerations}. \citet{mathew2019thou} show how this type of counterspeech is not well-accepted even by the communities in whose favour it is produced, and provide this example:

\begin{cs}{teal}{} 
\small CS: This is ridiculous!!!!!! I hate racist people!!!! Those police are a**holes!!!
\end{cs}

\noindent Similarly, \citet{benesch2016considerations} advise that \textit{Warning of consequences} should never turn into threats, as they show in this positive example:

\begin{cs}{teal}{} 
\small CS: Current and future employers will be able to see your tweets, using the hashtag created to attack the chancellor of your university, with misogynist and racist content.
\end{cs}

\noindent An empathetic, polite and constructive tone is also encouraged in guidelines written by counterspeech movements such as \textit{Get the Trolls out}.\footnote{\href{https://getthetrollsout.org/resources/online-hate-speech}{``Stopping hate: how to counter hate speech on Twitter''}.}

\noindent In \autoref{sec:classifying}, we discuss the task of automated identification of the strategies discussed here.\looseness=-1

\subsection{Related tasks} \label{sec:cn_related_tasks}

To better define counterspeech we describe its similarities and differences to several related tasks.\footnote{A non-exhaustive list of available datasets for these tasks can be found in Appendix \ref{app:related_datasets}.} The first of these is \textbf{hope speech}, which indicates comments with a constructive view of the future and a peace-seeking intent \cite{palakodety2019hope, chakravarthi2020hopeedi, kumaresan2023overview, garcia2023hope, jimenez2023overview}. However, as opposed to counterspeech, hope speech does not necessarily reply to hate speech or challenge a message with an opposing stance:

\begin{context}{gray}{} 
\small Context: Video about COVID-19.
\end{context}
\begin{reply}{gray}{} 
 \small Reply: We will survive these things. 
\end{reply}

An overlapping phenomenon with hate speech is online trolling: an aggressive online behaviour seeking strong reactions for the sake of amusement \cite{lee-EtAl:2022:LREC2}. Various \textbf{counter-trolling} strategies have been proposed. However, one of these consists in \textit{reciprocating}, while counterspeech writing guidelines discourage direct attacks on the interlocutor:

\begin{inputmess}{gray}{} 
\small Context: You gotta be a fucking idiot to be a sports fan lmao
\end{inputmess}

\begin{reply}{gray}{}
   \small Reply: Aren’t u that guy who has to give ur neighbors the biannual update that ur a sex offender
\end{reply} 

\citet{fraser2021understanding} instead, address stereotypes by generating \textbf{anti-stereotype} words rather than true counterspeech while \citet{allaway2023towards} provide counterstatements to essentialist beliefs (i.e. generalizations about a group):

\begin{inputmess}{gray}{} 
\small Context: I speak English, I don’t speak libt*rd
\end{inputmess}

\begin{reply}{gray}{} 
\small Reply: Conservatives can also be stupid.
\end{reply}

\noindent Even if counterspeech can contain anti-stereotyping messages, not all counterspeech strategies, such as asking \textit{clarifying questions} (e.g. \textit{What do you mean by ``libt*rd''?}), address the essentialist belief implicit in a text.

Other existing work has focused on making models better while handling dangerous or unsafe situations. \citet{kim2022prosocialdialog} present a dataset of \textbf{prosocial dialogues}, i.e. conversations in which the speaker responds to potentially unsafe situations by promoting respect for social norms. As opposed to counterspeech, these responses are directed to unsafe rather than hateful texts, as in this example:

\begin{inputmess}{gray}{} 
\small Context: I saw someone overdose and didn’t tell anyone.
\end{inputmess}

\begin{reply}{gray}{}
\small Reply: First of all, that must have been a terrifying situation. It was your responsibility to help the person in need though. If people find out that you saw the person overdose and didn't help, they will certainly not think highly of your behaviour either.
\end{reply}

Alternatively, \textbf{detoxification} \cite{laugier-etal-2021-civil, logacheva2022paradetox} consists in rephrasing toxic comments with civil wording: however, this does not address any possible underlying hateful content, as shown by the following example.

\begin{inputmess}{gray}{} 
\small Context: you now have to defend this clown along with his jewish corruption.
\end{inputmess}

\begin{reply}{gray}{}
\small Detoxified: you now have to defend this guy from his jewish ties.
\end{reply}

\citet{ung2022saferdialogues} make models open in receiving \textbf{feedback} from users \textbf{about safety failures} of their generated content. 
Although this feedback resembles \textit{denouncing} counterspeech, they tackle model-generated rather than user-generated hate speech:

\begin{inputmess}{gray}{} 
\small Context: I am getting a kick out of watching you try to think you have value in the family.
\end{inputmess}

\begin{reply}{gray}{}
\small Reply: no need to attack someone because you think differently.
\end{reply}

Another related task is \textbf{counter-argumentation} generation \cite{hua2018neural, hua2019sentence, hua-etal-2019-argument-generation, alshomary2021counter, alshomary2023conclusion}. Still, a logically valid counter-argument is not necessarily a good counterspeech, as shown in this example from \citet{fanton2021human}:

\begin{inputmess}{gray}{} 
\small Context: We should kill all the jews.
\end{inputmess}
\begin{reply}{gray}{} 
\small Reply: There are many alternatives to removing jews, such as converting them to another religion (e.g. Buddhism).
\end{reply}

Finally, \textbf{misinformation countering} consists of justifying the veracity of a statement \cite{stammbach2020fever, kotonya2020explainable, jolly2022generating, ma2023characterizing, he2023reinforcement, russo2023countering, russo2023benchmarking}.\footnote{We refer readers to \citet{he2023survey}'s survey, which analyses approaches to crowd-based and effective counter-misinformation.}
These justifications can have some  characteristics in common with counterspeech, e.g. being polite, fluent and relevant \cite{he2023reinforcement, russo2023countering}. 
However, counterspeech does not always contain evidence, and a factually inaccurate claim is not necessarily hateful, as shown in this example from \citet{russo2023countering}:

\begin{inputmess}{gray}{} 
\small Context: 11,000 of 13,000 knife attacks in London were carried out by Muslim migrants.
\end{inputmess}

\begin{reply}{gray}{}
\small Reply: This claim is baseless as information on offenders’ religion and nationality is not held by the authorities. Regardless, the claim is implausible.
\end{reply}

\section{Step 1: Design your task} \label{sec:task_design}
The first step is to select which counterspeech task(s) to tackle.
We discuss studies covering classification, selection and generation, and derive possible best practices from them.

\subsection{Classifying counterspeech} \label{sec:classifying}
Classification can help to understand counterspeech dynamics and to collect counterspeech data. We consider three sub-tasks.

\paragraph{CS detection.} Several works focus on detecting counterspeech as opposed to: non-counterspeech \cite{mathew2019thou, goffredo-etal-2022-counter, albanyan-etal-2023-finding}, hate speech \cite{garland2020countering}, hate speech and neutral instances  \cite{mohle-etal-2023-just, yu-etal-2022-hate, shah2022analysis, he2021racism, vidgen2021introducing}, and among Hostility, Criticism and Non-related instances \cite{vidgen2020detecting}. Finally, \citet{goffredo-etal-2022-counter} also identified messages supporting counterspeech.

\paragraph{User classification.} Only \citet{mathew2020interaction} worked on classifying Twitter users into hateful or counterspeakers: this task can be useful for a platform to intervene early and demote hateful accounts, while promoting counterspeech.

\paragraph{Strategy classification.} Detecting the counterspeech strategies\footnote{See \autoref{sec:taxonomies} for an overview of counterspeech strategy taxonomies.} used \cite{mathew2019thou, chung2021multilingual} can help to analyse their effectiveness and develop more fine-grained responses.
Similarly, \citet{albanyan2022pinpointing} identify counterspeech, determining whether it provided a justification, attacks the author of the hate speech, or includes additional hate. \\

\noindent While some of these classification studies employ only traditional classifiers \cite{mathew2020interaction, shah2022analysis}, others compare them with neural models, showing that the latter perform better or comparably well than the first \cite{mathew2019thou, he2021racism, vidgen2021introducing}. Most studies employ only neural models and experiment with different types of input, showing how including the context (e.g. the hate speech) helps to reduce false negatives~\cite{vidgen2021introducing, yu-etal-2022-hate, albanyan2022pinpointing}. 
In fact, hate speech and counterspeech can share similar textual features to some extent, making it difficult to automatically distinguish them without further context~\cite{mohle-etal-2023-just}. Better counterspeech detection performance is obtained by pretraining the models on similar tasks, such as stance~\cite{yu-etal-2022-hate} or emotion detection~\cite{albanyan2022pinpointing}.

\begin{protip}{Goldenrod!10!white}{}
\includegraphics[scale=0.03]{wrench.png} 
The most common errors in counterspeech classification arise when the text is complex and contains irony or sarcasm \cite{goffredo-etal-2022-counter, albanyan2022pinpointing}, negation \cite{yu-etal-2022-hate}, or when more context is needed to disentagle counterspeech from other categories \cite{vidgen2021introducing}. Another problem, common to hate speech detection, is lexical overfitting to specific terms or swearwords \cite{vidgen2020detecting, yu-etal-2022-hate}.  In other cases, errors might arise from the annotation itself \cite{vidgen2020detecting}. Using a large enough dataset with high-quality annotation can help to reduce such errors.
\end{protip}

\subsection{Selecting counterspeech responses}
One way to produce counterspeech consists of selecting from a pool of possible responses that can be obtained via over-generation \cite{zhu-bhat-2021-generate}. Alternatively, the candidates can be retrieved from a counterspeech dataset: \citet{chung2021empowering} rely on a \textit{tf-idf} information retrieval model, while \citet{akazawa-etal-2023-distilling} employ the implicit stereotype of the hate speech to make a selection via cosine similarity. It is also possible to select counterspeech among non-counterspeech content available online, e.g. from Twitter \cite{mohle-etal-2023-just} or online articles~\cite{albanyan-etal-2023-finding}.

\begin{protip}{Goldenrod!10!white}{}
\includegraphics[scale=0.03]{wrench.png} 
Filtering a social media dataset containing both counterspeech and non-counterspeech instances does not produce a larger amount of counterspeech than a random sample~\cite{mohle-etal-2023-just}. Thus, selection seems particularly useful to obtain the most appropriate response to a specific hate speech when a pool of gold \cite{akazawa-etal-2023-distilling, chung2021empowering} or silver \cite{zhu-bhat-2021-generate} counterspeech is already available, rather than filtering out non-counterspeech instances. 
\end{protip}

\subsection{Generating counterspeech}
Suitable counterspeech can take many forms: we outline non-exhaustive desirable aspects of counterspeech (knowledge, personality, style), and report relevant techniques for generation (fine-tuning and prompting, translation). 

\paragraph{Knowledge guided generation.} 
Both \citet{chung-etal-2021-towards} and \citet{jiang2023raucg} structure this task in two phases: first the extraction of relevant knowledge from an external source, and secondly the generation of knowledge-augmented counterspeech. For the first phase, \citet{chung-etal-2021-towards} used extracted keyphrases to select sentences from Wikipedia articles and news datasets, while \citet{jiang2023raucg} rely on stance consistency, semantic overlap rate, and fitness for hate speech to construct a knowledge repository from the ChangeMyView subreddit.

\paragraph{Personality guided generation.} 
Examples of this approach are \citet{de2021toxicbot}, who employed the PersonaChat dataset to fine-tune a model provided with a dynamic persona profile or dialogue history as input during generation, and \citet{doganc-markov-2023-generic}, who experimented with both fine-tuning and few-shot prompting to incorporate the profiling information and obtain personalized counterspeech.

\paragraph{Style guided generation. }
Here, we include all other stylistic features addressed during generation. To enhance \textit{specificity}, \citet{bonaldi-etal-2023-weigh} employ two attention-based regularization techniques to include a broader context during training and generation, while \citet{furman-etal-2023-high} focus on the \textit{argumentative information} present in the hate speech to guide the generation towards particular response strategies.
Other works target multiple aspects at the same time: 
\citet{ijcai2022p716} simultaneously control for the \textit{politeness}, \textit{detoxification} and \textit{emotion} in the generated counterspeech. Finally, \citet{gupta-etal-2023-counterspeeches} propose a two stage-framework for generating counterspeech conditioned on five different \textit{strategies} (i.e. informative, denouncing, question, positive, and humour).

\begin{protip}{Goldenrod!10!white}{}
\includegraphics[scale=0.03]{wrench.png} 
Despite the importance of knowledge-driven generation, correcting misinformation alone is not sufficient and can lead to higher levels of violence \cite{carthy2023countering}. For this reason, taking into consideration other aspects is fundamental: for example, \citet{hangartner2021empathy} showed how empathy-based counterspeech can have an impact, however small, in reducing hate speech. Moreover, generating counterspeech with specific strategies according to the targeted community can be particularly effective, and in general, maintaining a polite tone is recommended~\cite{mathew2019thou}.
\end{protip}

\paragraph{Fine-tuning and prompting.}
The most commonly employed approach for counterspeech generation is fine-tuning a language model on a counterspeech dataset \cite[e.g.][]{qian-etal-2019-benchmark, tekiroglu2022using, halim2023wokegpt}. 
However, recent advances have allowed generation of counterspeech via few-shot \cite{ashida2022towards, furman-etal-2023-high, vallecillo2023automatic, doganc-markov-2023-generic}, one- and zero-shot prompting \cite{munbeyond, zheng-etal-2023-makes}. 

\paragraph{Translation and low-resourced languages.}
\citet{chung2020italian} generate Italian counterspeech by fine-tuning a model on a combination of gold and silver Italian data obtained via translation. Also \citet{vallecillo2023automatic} rely on translated examples from \citet{chung-etal-2021-towards} to create a Spanish corpus via few-shot prompting. 
Finally, \citet{furman-etal-2023-high} include Spanish examples in their generation task.

\begin{protip}{Goldenrod!10!white}{}
\includegraphics[scale=0.03]{wrench.png} 
Prompting allows generation of counterspeech in a low computationally intensive way: however, given the specificity of the task, few-shot prompting is preferred over one- and zero-shot prompting. Moreover, clear and specific instructions should be given to the model to obtain more fine-grained replies. In particular, both \citet{hassan-alikhani:2023:ijcnlp} and \citet{munbeyond} show how LLMs tend to use general strategies such as \textit{denouncing}, \textit{comment} or \textit{correction} when generating counterspeech without specific indications. 
Another viable strategy to obtain data in low-resourced scenarios is translation, as shown by \citet{chung2020italian} and \citet{vallecillo2023automatic}, who respectively use silver translated data alone or together with gold data in the language of interest to generate responses in Spanish and Italian.
\end{protip}

\section{Step 2: Select the data} \label{sec:select_data}
After task design, the next choice is whether to collect a new dataset or to use an already existing one. We will discuss the use-cases of the main counterspeech collection procedures, and then detail the characteristics of available counterspeech datasets.

\subsection{Collecting your own data}
Collecting data entails specific consequences according to the chosen strategy \cite{tekiroglu-etal-2020-generating}: we summarise them below and in Table \ref{tab:data_collection_procedures}.

\begin{table}[htbp]
\small
\centering
\resizebox{\columnwidth}{!}{
\begin{tabular}{llcccc} 
\toprule
\multicolumn{1}{l}{\textbf{Coll.}}
                  & \multicolumn{1}{l}{\textbf{Data type}} & \multicolumn{1}{c}{\textbf{Quant.}} & \multicolumn{1}{c}{\textbf{Conf.}} & \multicolumn{1}{c}{\textbf{Div.}} & \multicolumn{1}{c}{\textbf{Non-eph.}} \\ \midrule
Crawl. & Real & \ding{51} & - & \ding{51} & - \\
Crowd. & Simulated & \ding{51} & \ding{51} & - & \ding{51} \\
Niche. & Simulated & - & \ding{51} & \ding{51} & \ding{51} \\
Hybr. & Synthetic & \ding{51} & \ding{51} & \ding{51} & \ding{51}\\ 
Auto. & Synthetic & \ding{51} & \ding{51} & - & \ding{51} \\
\bottomrule
\end{tabular}}
    \caption{Data type, quantity, conformity to counterspeech writing guidelines, diversity and non-ephemerality of counterspeech collected with different procedures.}
    \label{tab:data_collection_procedures}
\end{table}

\paragraph{Crawling.} This consists of scraping real counterspeech from sources such as Youtube \cite{mathew2019thou}, Twitter \citep[e.g.][]{mathew2020interaction, vidgen2020detecting, goffredo-etal-2022-counter}, Reddit \cite{yu-etal-2022-hate, vidgen2021introducing, hassan-alikhani:2023:ijcnlp}, and online articles \cite{albanyan-etal-2023-finding}. Crawling can guarantee high quantity and diversity of data. However, responses in the wild do not always conform to counterspeech guidelines, as shown in this example from \citet{albanyan2022pinpointing}:

\begin{hs}{carmine}{} 
\small HS: on my way to f**k your b**ch.
\end{hs}

\begin{csreply}{teal}{}
\small CS: But my b**ch is your mamma
\end{csreply}

To tackle such problems, \citet{hassan-alikhani:2023:ijcnlp} manually remove offensiveness and first-person references from their crawled data. Furthermore, relying on lists of keywords or users may not capture the entirety of available counterspeech. Finally, crawled data might be ephemeral, hindering reproducibility\footnote{
\citet{albanyan2022pinpointing} and \citet{albanyan-etal-2023-counterhate} include only the Tweets IDs and not the full texts.} 
\cite{klubicka2018examining}.

\paragraph{Crowdsourcing.} \citet{qian-etal-2019-benchmark} and \citet{furman2022parsimonious, furman-etal-2023-high} rely on non-expert annotators to manually produce counterspeech. With crowdsourcing (Crowd.) it is possible to obtain a high quantity of data, although these often contain simple and poorly varied arguments. For example, a large amount of human-written counterspeech can consist in \textit{denouncing} the hate speech: even if this is an acceptable strategy, it represents a simple form of reply, as shown in this example from \citet{qian-etal-2019-benchmark}:

\begin{hs}{carmine}{} 
\small HS: Like... I literally addressed this point and you completely ignore it, like a stupid retard nigger.
\end{hs}

\begin{csreply}{teal}{}
\small CS: Stop using racist and medical conditions as an insult.
\end{csreply}

\paragraph{Nichesourcing.} This relies on experts to produce counterspeech
\cite{conan-2019, chung-etal-2021-towards}. Nichesourcing (Niche.) can guarantee the highest quality data in terms of syntactic variety and semantic complexity. However, it is also the most expensive strategy, making it difficult to obtain a large quantity of data.

\paragraph{Hybrid and fully automated collection.}
Hybrid approaches (Hybr.) exploit a combination of automatic and manual intervention to collect data. \citet{tekiroglu-etal-2020-generating} propose an author-reviewer architecture where a generation model (author) produces hate speech and counterspeech data, and human reviewers filter and post-edit the generated data. This strategy is then employed by both \citet{fanton2021human} and \citet{bonaldi-etal-2022-human}. Below, an example of counterspeech post-editing from \citet{fanton2021human} is shown:

\begin{hs}{carmine}{} 
\small HS: Transgenders want to rape our children.
\end{hs}
\begin{replyx}{gray}{}
\small CS: This is not true. Maybe they are worried because of the rise in hate crimes, incidents of which are down to 28 percent, since 2014. 
\end{replyx}
 \begin{csreply}{teal}{}
\small CS$_{pe}$: This is not true. Maybe \textit{you should be worried} about the rise in hate crimes \textit{against queers}, incidents of which \textit{are almost doubled} since 2014.
\end{csreply}

\citet{ashida2022towards} and \citet{vallecillo2023automatic} fully rely on generative models to produce counterspeech (Auto.): human annotators do not intervene.

\begin{table*}[htbp]
\small
\centering
\begin{tabularx}{\textwidth}{lrrllXlcc}
\toprule
\textbf{Dataset}                 & \textbf{Size} & \textbf{\# CS} & \textbf{Interact.}   & \textbf{Coll.} & \textbf{Source}  & \textbf{Lang.}     & \textbf{Tar.} & \textbf{Add.} \\
\midrule

\citet{mathew2019thou}           & 13,924         & 6,898         & Pairs + c. & Crawl.   & YouTube & EN & \ding{51} & \ding{51} \\
\citet{conan-2019}               & 14,988         & 14,988         & Pairs & Nich. & NGOs op. & EN/FR/IT  & \ding{51} & \ding{51} \\ 
\citet{qian-etal-2019-benchmark} & 16,845         &  29,388     & Pairs + c. & Crowd. & Reddit, Gab & EN  & - & - \\ 
\citet{mathew2020interaction}    & 1,290          & 1,290          & Pairs & Crawl. & Twitter  & EN & - & \ding{51} \\
\citet{vidgen2020detecting}      & 20,000         & 116           & Single c. & Crawl. & Twitter  & EN & \ding{51} & - \\ 
\citet{he2021racism}             & 2,290          & 517           & Single c. & Crawl. & Twitter  & EN & \ding{51} & \ding{51} \\ 
\citet{vidgen2021introducing}    & 27,494         & 220           & Single c. & Crawl. & Reddit  & EN & - & - \\ 
\citet{chung-etal-2021-towards}  & 195           & 195           & Pairs & Niches. & NGO op.    & EN & \ding{51} & \ding{51} \\
\citet{fanton2021human}          & 5,003          & 5,003          & Pairs & Hybr.        & NGOs op.  & EN & \ding{51} & - \\ 
\citet{yu-etal-2022-hate}        & 6,846          & 1,622              & Pairs & Crawl. & Reddit  & EN & - & \ding{51}  \\
\citet{albanyan2022pinpointing}  & 5,652          & 1,149          & Pairs & Crawl. & Twitter  & EN & - & \ding{51} \\
\citet{bonaldi-etal-2022-human}         & 3,059           &  8,311     & Dialog. & Hybr.        & NGOs op.  & EN & \ding{51}  & - \\   
\citet{ashida2022towards}         & 348 &    306           & Pairs & Autom.        & Autom.  & EN & - & \ding{51}  \\  
\citet{goffredo-etal-2022-counter} & 624 & 81 & Pairs & Crawl. & Twitter & IT & \ding{51} & \ding{51}  \\
\citet{furman2022parsimonious}   & 2,055         & 2,055             & Pairs & Crowd. & \citet{basile2019semeval} & ES & - & \ding{51}  \\
\citet{furman-etal-2023-high}        & 2,077         & 2,077          & Pairs & Crowd. & \citet{furman2023which} & EN/ES & - & - \\
\citet{vallecillo2023automatic} & 238 & 238 & Pairs & Autom. & \citet{chung-etal-2021-towards}  & ES & \ding{51} & \ding{51}  \\
\citet{hassan-alikhani:2023:ijcnlp} & 3,900 & 250 & Pairs & Crawl. & Reddit & EN & \ding{51} & \ding{51} \\
\citet{albanyan-etal-2023-counterhate} & 2,621 & 1,685 & Pairs + c. & Crawl. & Twitter & EN & - & \ding{51} \\
\citet{albanyan-etal-2023-finding} & 54,816 & 2,365 & Pairs & Crawl. & Web articles & EN & \ding{51} & - \\
\bottomrule
\end{tabularx} 
\caption{Available datasets, according to their size, nr. of counterspeech interaction type, data collection procedure, source, language, target, and additional information. The data size and the number of counterspeech refer to the interactions shape (e.g. 5,003 \textit{pairs}), except for \citet{qian-etal-2019-benchmark} and \citet{bonaldi2022human} where the number of effective counterspeech turns is shown.}
\label{tab:datasets}
\end{table*}

\begin{protip}{Goldenrod!10!white}{}
\includegraphics[scale=0.03]{wrench.png} 
\textit{Crawling} is the most common data collection procedure used in the wild to gather counterspeech. 
However, \textit{nichesourcing} can generate the highest-quality responses, since it benefits form expert knowledge. 
If expertise is limited, partially automatising data collection via a \textit{combination} of a fine-tuned model and human post-editing can be a good solution. If expertise is extremely limited, non-expert annotators or a classifier \cite{hassan-alikhani:2023:ijcnlp} can prefilter the data prior to expert validation~\cite{tekiroglu-etal-2020-generating}. Alternatively, non-expert annotators can be trained, following the procedure described in Appendix \ref{app:annot_training}. However, we discourage relying solely on automatic 
counterspeech collection without human intervention, given the sensitivity of this task.
\end{protip}

\subsection{Choosing from existing datasets}
An efficient alternative to data collection is selecting among available counterspeech datasets. 
We describe them along several dimensions, summarised in Table \ref{tab:datasets}, to facilitate the choice of most suitable dataset for specific research needs.

\paragraph{Shape of the interactions.} 
Available datasets can be divided into four main groups according to the type of interaction they contain. Single comments (Single c.) are individually labeled as hate speech, counterspeech, or other classes, without further conversational context, and often come from social media platforms such as Twitter or Reddit~\cite{vidgen2020detecting, vidgen2021introducing,he2021racism}. Pairs of hate speech and their related counterspeech are the most widely diffused type of interaction encoded in available datasets \cite[e.g.][]{conan-2019, goffredo-etal-2022-counter, vallecillo2023automatic}. 
Alternatively, pairs with context (Pairs+c.) include a longer conversational context,
such as previous or subsequent comments \cite{mathew2019thou, qian-etal-2019-benchmark,albanyan-etal-2023-counterhate}.
Finally, \citet{bonaldi-etal-2022-human} present hate speech and counterspeech dialogues (Dialog.) including multiple counterspeech turns.

\paragraph{Targets of hate.} 
Most studies include multiple targeted minorities: the most represented are \textit{Jews}, \textit{Blacks}, and \textit{LGBT} \cite{mathew2019thou}. Additionally, \citet{chung-etal-2021-towards}, \citet{bonaldi-etal-2022-human} and \citet{vallecillo2023automatic} consider \textit{Migrants}, \textit{Muslims},  and \textit{Women}, \citet{hassan-alikhani:2023:ijcnlp} cover \textit{Disabled} people, and \citet{fanton2021human} include \textit{Overweight} and \textit{Romani} people on top of these. Other studies focus on a single target: in particular, \citet{conan-2019} on \textit{Islamophobia}, whereas \citet{he2021racism} and \citet{vidgen2020detecting} on COVID-19 related \textit{Asian} hate. 
Only \citet{fanton2021human} include, in a few examples, intersectional hate, i.e. hate directed towards people belonging to multiple minorities, such as \textit{black women}. Finally, \citet{albanyan-etal-2023-finding} is the only research to address hate towards \textit{individuals}, rather than groups.

\paragraph{Types of hate addressed.} 
\citet{conan-2019} identify hate speech according to the sub-topic it covers: \textit{culture}, \textit{economics}, \textit{crimes}, \textit{rapism}, \textit{women oppression}, \textit{history} and \textit{other/generic}. \citet{vidgen2020detecting} make different levels of distinctions according to the offensiveness intensity (\textit{Hostility} and \textit{Criticism}) and category (\textit{Interpersonal abuse}, \textit{Use of threatening language} or \textit{Dehumanization}). Finally, \citet{vidgen2021introducing} distinguish hate according to the addressed entity (an \textit{identity}, an \textit{affiliation} or an \textit{identifiable person}) and category (\textit{derogation}, \textit{animosity}, \textit{threatening}, \textit{dehumanization} or \textit{glorification of hateful entities}).

\paragraph{Languages.} 
Most existing datasets are in English, with only a few covering French \citep{conan-2019}, Italian \citep{conan-2019, goffredo-etal-2022-counter}, and Spanish \citep{furman2022parsimonious, furman-etal-2023-high, vallecillo2023automatic}. 

\paragraph{Additional information.} 
Other information present in these datasets include the counterspeech strategy \citep[][see Section \ref{sec:taxonomies}]{mathew2019thou, conan-2019, mathew2020interaction, goffredo-etal-2022-counter}.  
Similarly, \citet{albanyan2022pinpointing} specify whether the counterspeech contains a justification or attacks the author, and if it is not a counterspeech whether it agrees with the hater or adds additional hate. \citet{albanyan-etal-2023-counterhate} do the same but for replies to counterspeech. Others have considered the \textit{discourse}\footnote{From the Segmented Discourse Representation Theory \cite{asher2003logics}} \cite{hassan-alikhani:2023:ijcnlp} and the  \textit{argumentative strategy} countering the hate speech \cite{furman2022parsimonious, furman-etal-2023-high}.  

Then, there can be contextual information on social media platforms data, such as the title of the discussion \cite{yu-etal-2022-hate}, the list of replies and the timestamp \cite{mathew2019thou}, the number of likes and replies \cite{mathew2019thou}, or the ego network of the users \cite{he2021racism}.

Other aspects are the annotator demographics, \cite{conan-2019}, or the human annotations: \citet{ashida2022towards} and \citet{vallecillo2023automatic} include the counterspeech offensiveness, stance and informativeness. Further fine-grained information include knowledge sentences \citep{chung-etal-2021-towards} and paraphrases of crawled counterspeech via the removal of offensiveness and first-person references \cite{hassan-alikhani:2023:ijcnlp}.

\begin{protip}{Goldenrod!10!white}{}
\includegraphics[scale=0.03]{wrench.png} 
The choice of the dataset should be driven by task design. It is important to consider the dataset size and the actual number of counterspeech instances: a few examples can be enough for few-shot prompting, while larger datasets are beneficial for fine-tuning or selection tasks. 
Additionally, the source and procedure of data collection can affect the structure, style, and strategies of the included counterspeech (e.g. Tweets are shorter, crowdsourced data often contain \emph{denouncing} counterspeech). Finally, any additional information can be decisive according to the specific goal of the study, e.g. the knowledge sentences in the dataset by \citet{conan-2019} for knowledge-driven generation.
\end{protip}

\section{Step 3: Evaluate} \label{sec:evalution}
Next, we look at the literature on evaluating counterspeech based on the tasks discussed in \S\ref{sec:task_design}.\footnote{The metrics described in \S\ref{sec:eval_gen} are meant for evaluating generation tasks, but they can be used for selection too.}

\subsection{Evaluating classification}
When gold test data is available, performance can be assessed via F1, precision, recall, accuracy, and confusion matrices.
Moreover, in multi-label scenarios (e.g. counterspeech employing multiple strategies), hamming loss is recommended as it can better capture model performance by considering the ratio of true classes in a prediction rather than a hard right prediction \citep{mathew2019thou}. 
Finally, human judgement can be compared to classifiers to verify their performance \citep{garland2020countering}, and qualitative error analysis can help to better understand the specific flaws of a model \cite{vidgen2020detecting, vidgen2021introducing, goffredo-etal-2022-counter, yu-etal-2022-hate}. 

\subsection{Evaluating generation} \label{sec:eval_gen}
Standard evaluation metrics can be grouped into extrinsic (measuring the potential impact of a system on its related tasks or on achieving its overall goals) and intrinsic measures \citep[assessing the system output in isolation,][]{walter-1998-book}.

\paragraph{Extrinsic evaluation.} 
So far, only \citet{chung2021empowering} have focused on this kind of evaluation.
To assess how effective their counterspeech suggestion tool was in empowering NGO operators during hate countering, the operators were asked to evaluate their user experience through a questionnaire \cite{laugwitz2008construction} and open-ended qualitative questions.\footnote{For a detailed discussion on evaluating the impact of counterspeech in real-life scenarios, see
\citet{chung2023understanding}.} 

\paragraph{Intrinsic automatic metrics.} Some of these metrics centre on the comparison between generation and references using criteria such as linguistic surface \cite{papineni2002bleu, lin2004rouge}, novelty \cite{wang2018sentigan}, and semantic similarity \cite{zhang2019bertscore}. Furthermore, some work measures the quality of counterspeech generation based on fine-grained characteristics, such as toxicity \cite{perspectiveAPI}, informativeness \cite{fu2023gptscore}, factuality \cite{fu2023gptscore}, repetitiveness \cite{bertoldi2013cache, cettolo2014repetition}, linguistic acceptability, politeness, emotion \cite{ijcai2022p716}, stance and relevance to the input \citep[i.e. the hate speech,][]{schutze2008introduction, halim2023wokegpt}. 

\paragraph{Human evaluation.} 
Several factors should be considered, such as evaluation criteria, scale (e.g. ranking vs. Likert or sliding scale), and annotators (e.g. experts vs. crowd). The common approach is to ask annotators to judge responses on a scale (e.g. of 1 to 5) based on aspects including suitableness and specificity \cite{chung-etal-2021-towards, tekiroglu-etal-2022-using, bonaldi-etal-2023-weigh}, grammaticality \cite{chung2020italian, zhu-bhat-2021-generate}, coherence and informativeness \cite{chung-etal-2021-towards}.

\begin{protip}{Goldenrod!10!white}{}
\includegraphics[scale=0.03]{wrench.png} 
While intrinsic automatic metrics can capture the overall performance of generation systems at scale, some of these lack interpretability and correlation with human evaluation \cite{belz2006comparing, novikova2017we}. 
Considering the complexity of hate mitigation, human evaluation is a more reliable approach. Most previous work uses experts or trained annotators for manual evaluation. Since the choice of the best response is subjective, it is desirable to enlist diverse annotators \citep[e.g. in regard to gender and educational level,][]{waseem2017understanding, sap2019risk,abercrombie-etal-2023-resources} or users identifying with the potential recipients of counterspeech such as perpetrators and bystanders.
\end{protip}

\section{Open challenges} 

Drawing from the surveyed literature, we highlight key open challenges in counterspeech research.

\paragraph{Language and culture.} 
Hate speech is not only linguistically, but also culturally specific. Therefore, it requires culturally specific responses. 
For example, in Spanish, the same words can convey discriminatory connotations depending on the country in which they are used \cite{castillo2023analyzing}. Moreover, the same groups can be subject to different stereotypes associated with the historical events of their location \cite{laurent2020project}.

\paragraph{Sources of hate.} A level of granularity not yet considered for counterspeech design is the identity of the hate speech perpetrator. 
This, in turn, can be considered together with cultural and geographical factors, to produce counterspeech tailored to specific targets (e.g. Italian neonazis).

\paragraph{Types of hate.} Studies on counterspeech are mostly centred on explicit hate with only a few addressing stereotypes, prejudice or biases \cite{munbeyond}. Such implicit hate often contains complex linguistic forms with indirect sarcasm or humour \cite{waseem2016hateful, nunes2018survey}, and can be generic \citep[``\textit{boys play with trucks}'',][]{doi:10.1073/pnas.1208951109, leslie2014carving}, posing challenges in how to mitigate it~\cite{buerger2022they}.

\paragraph{Hallucinations.} 
Even if counterspeech does not necessarily need to contain factual evidence (\S\ref{sec:cn_related_tasks}), it can be effective in highlighting the absurdity of hate speech.
However, a challenge in open-ended generation is hallucinations.\footnote{I.e, text nonsensical or unfaithful to the provided source input \cite{10.1145/3571730}, often with factually incorrect content.}
One way to address this is to rely on external knowledge sources \cite{chung-etal-2021-towards, jiang2023raucg}: here, RAG systems \cite{lewis2020retrieval,ram2023context} are a promising research direction.
Alternatively, inaccurate text can be detected in the generation \cite{manakul2023selfcheckgpt}.
Finally, counterspeech should be placed in the right temporal context to be more effective: knowledge-grounded generation can help to produce more time-relevant responses.

\paragraph{Evaluation.} As discussed in Section \ref{sec:evalution}, existing evaluation metrics are limited. It would be desirable to create test suites analysing different functionalities of counterspeech generation models; e.g. testing models' capacity to generate counterspeech directed at specific types of hate with certain strategies \citep[similar to the HateCheck initiative, ][]{rottger-etal-2021-hatecheck}. Additionally, the definition of good counterspeech is subjective and should be user-oriented (e.g. assessed by the target audience). Hence, an ideal evaluation could involve gathering multiple perspectives on suitable counterspeech.

\paragraph{Biases in data collection.} Possible biases can emerge from various choices taken during data collection. 
Firstly, the data source can strongly affect content and style. With crawling, collecting texts from a specific platform will determine its length, style and topics, mainly representing the users of that platform and thus not being highly generalisable. 
With crowdsourcing and nichesourcing, it should be considered that annotators have different sensitivity to hate, according to their country of origin \cite{lee2023crehate}, their belonging to a targeted minority, and their personal experiences. This can have a considerable impact on the content of the counterspeech they write too. Moreover, non-experts recur more often to simpler counterspeech strategies such as \textit{denouncing} than experts \cite{tekiroglu-etal-2020-generating}. 
Choice of annotators also creates similar possible biases to human evaluation, as discussed above and in section \ref{sec:eval_gen}: for all these reasons, it is better to recruit a diverse set of annotators, if possible.
Finally, bias can also originate from the other factors already discussed, i.e. the considered targets of hate, language and geographical/temporal context of the collected data. In general, it is always preferable to provide newly introduced datasets with a dataset card\footnote{\url{https://huggingface.co/docs/hub/datasets-cards}} to inform users on how to responsibly employ the data, limiting the emergence of possible harms.

\section{Conclusion}
We presented a thorough review of 43 NLP studies on counterspeech. This is organised as a step-by-step guide, intended for those approaching counterspeech from an NLP perspective. First, we framed counterspeech and its strategies, distinguishing it from other similar tasks. Then, we structured the subsequent sections as progressive steps to undertake when approaching counterspeech research in NLP: in these sections, we relied on the literature to provide insights into the consequences each choice might imply. Finally, we point out open challenges in the field. Counterspeech represents a promising approach to tackling online hate, and NLP can potentially provide the tools to make it scalable. However, an efficient system is not necessarily a good system: researchers operating in this area must be aware of the consequences entailed by each of their choices, to avoid spreading further harm.

\newpage

\section*{Limitations}
To an external reader, the number of papers included in this study might seem small: this is both because relatively little attention has still been devoted to this topic, and because we made the specific choice of focusing only on NLP papers proposing one or more of the following contributions: a dataset, a classification, selection or generation task. 
In this survey, we included studies from Scopus, arXiv and the ACL Anthology, following the methodology of previous abusive language surveys in the NLP domain \cite{chung2023understanding, vidgen2020directions}. Our search was conducted using keywords, which might not be comprehensive of all the available studies on counterspeech but represents a reasonable compromise for searching in such huge databases. Moreover, all the authors already had research experience in counterspeech and thus had a personal list of counterspeech studies collected over the years---all of which were retrieved with the automated search process.

\section*{Ethical considerations}
In addition to potential legal issues \citep[see][for a discussion of this]{chung2023understanding}, engaging in counterspeech has important social consequences: for this reason, many precautions should be adopted when dealing with it, similarly to other abusive language related domains. First of all, researchers and potential annotators involved in any counterspeech task should prioritise their mental well-being: prolonged exposure to abusive content can have negative effects, that can be avoided by following the mitigation measures described by \citet{vidgen-etal-2019-challenges} and \citet{kirk-etal-2022-handling} (see Appendix \ref{app:annot_training} for more details). 
For what regards data collection and distribution, synthetic data represent a viable option to preserve users privacy. Moreover, using simulated hate speech that are simple and stereotyped can avoid possible negative outcomes such as training a language model on hate speech generation. However, if the collected data are real, it is important to ensure that this does not 
interfere with the online activities of counterspeakers. For example, if the counterspeech included in a dataset are obtained by scraping from a list of activists' accounts, malicious users might reverse-search these texts, and identify the operators’ accounts, thus exposing them to possible attacks.
Finally, regarding the deployment of generation systems in real-life scenarios, human supervision is still necessary: the risks of hallucinations and abusive generation are still too high to fully automate the task of counterspeech production in the wild.

\section*{Acknowledgements}

Gavin Abercrombie was supported by the EPSRC project `Equally Safe Online' (EP/W025493/1). Yi-Ling Chung was supported by the Ecosystem Leadership Award under the EPSRC Grant EPX03870X1 \& The Alan Turing Institute.

\clearpage

\bibliography{anthology,custom}

\clearpage
\appendix
\section{Appendix}

\subsection{Methodology of the review }\label{app:meth_review}

\begin{figure}
    \centering
    \includegraphics[width=1\linewidth]{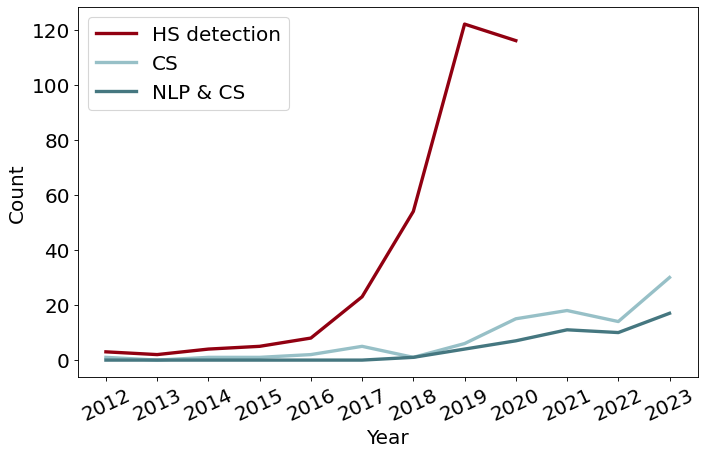}
    \caption{Number of published papers about hate speech detection, counterspeech in general and in the field of NLP. Data for hate speech detection cover only the 2013-2020 time span \cite{jahan2023systematic}.}
    \label{fig:papers_statistics}
\end{figure}

Figure \ref{fig:papers_statistics} shows how the number of published counterspeech papers is subject to a steady growth. 
To select the studies reviewed in this survey, we follow the PRISMA framework \cite{moher2010preferred}. The main aim of this review is to provide a guide for tackling counterspeech tasks in the NLP area. Therefore, as inclusion criteria, we only include publicly available papers presenting (i) a computational approach to (ii) text-based tasks that are (iii) related to online counterspeech. In particular, all the included papers either present a data collection, or concern classification, generation or selection tasks. Following previous reviews on counterspeech and abusive language \cite{vidgen2020directions, chung2023understanding}, we used three different sources to select the publications of our interest: the ACL Antology, Scopus and arXiv. First, we searched in these databases for all the publications including at least one of the following keywords: \textit{counterspeech}, \textit{counter-speech}, \textit{counter speech}, \textit{counter narratives}, \textit{counter-narratives}, \textit{counter hate}, \textit{counter-hate}, \textit{counterhate}, \textit{hate countering}, \textit{countering online hate speech}. Similarly to \citet{vidgen2020directions}, since Scopus includes a much broader content, we limited the subject area to \textit{Computer Science}. The automatic selection resulted in 156 papers from Scopus, 31 from arXiv and 20 from the ACL anthology\footnote{The search was last implemented on 14 December 2023.}. 23 duplicates were removed. Then, two of the authors manually revised the automatically filtered publications: first they considered only those which were NLP-related (shown as \textit{NLP \& CS} in Figure \ref{fig:papers_statistics}). Then, from this subset, they kept only the publications that either presented a data collection, generation, classification or selection task. They also removed too short (e.g. 2 pages initial studies) or not publicly available papers. The disagreements between the authors (regarding 3 papers) were solved by discussion. After this filtering, a total of 43 papers were included in this survey. 

\subsection{Examples of counterspeech taxonomies} \label{app:taxonomies_table}
\begin{table*}[htbp]
\small
\centering
\begin{tabularx}{\linewidth}{ p{0.1\linewidth} p{0.2\linewidth} X }
\toprule
& \textbf{Strategy}                  & \textbf{Example}       \\
\midrule 
\multirow{23}{\hsize}{\citet{benesch2016}}  & \textbf{Presenting facts} & Actually homosexuality is natural. Nearly all known species of animal have their gay communities. Whether it be a lion or a whale, they have or had (if they are endangered) a gay community.\\ \addlinespace 
& \textbf{Pointing out hypocrisy} & The ‘US Pastor’ can’t accept gays because the Bible says not to be gay. But...he ignores: The thing about eating shrimp or pork, [...] The thing about working on the Holy Day (Saturday or Sunday depending)...for any and all of those sins one should burn for an eternity, yet is ignored. \\ \addlinespace 
& \textbf{Warning of consequences} & I’m not gay but nevertheless, whether You are beating up someone gay or straight, it is still an assault and by all means, this preacher should be arrested for sexual harassment and instigating!!! \\ \addlinespace 
& \textbf{Affiliation} & Hey I’m Christian and I’m gay and this guy is so wrong. Stop the justification and start the accepting. I know who my heart and soul belong to and that’s with God: creator of heaven and earth. We all live in his plane of consciousness so it’s time we started accepting one another. That’s all. \\ \addlinespace 
& \textbf{Denouncing hateful speech} & please take this down YouTube. this is hate speech. \\  \addlinespace 
& \textbf{Humor and sarcasm} & Of course Jews are focused on 'world domination', even "galaxy domination". But so are Sith Order, Sauron etc.
\\ 
\midrule 
\multirow{6}{\hsize}{\citet{mathew2019thou}} & \textbf{Positive tone} & I am a Christian, and I believe we’re to love everyone!! No matter age, race, religion, sex, size, disorder...whatever!! I LOVE PEOPLE!! We are not going to go anywhere as a country if we don’t put God first in our lives, and treat EVERYONE with respect. \\ \addlinespace 
& \textbf{Hostile language} & This is ridiculous!!!!!! I hate racist people!!!! Those police are a**holes!!! \\ 
\midrule 
\citet{conan-2019} & \textbf{Counter-questions} & Is this true? Where is your source? \\
\midrule
\multirow{10}{\hsize}{\citet{qian-etal-2019-benchmark} } & \textbf{Identify Hate Keywords} & The C word and language attacking gender is unacceptable. Please refrain from future use. \\ \addlinespace 
& \textbf{Categorize Hate Speech} & The term fa**ot comprises homophobic hate, and as such is not permitted here. \\ \addlinespace 
&\textbf{Positive Tone Followed by Transitions} & I  understand your frustration, but the term you have used is offensive towards the disabled community. Please be more aware of your words. \\ \addlinespace 
& \textbf{Suggest Proper Actions} & I think that you should do more research on how resources are allocated in this country. \\ 
\midrule 
\multirow{8}{\hsize}{\citet{vidgen2020detecting}}& \textbf{Reject the premise of abuse} & it isn’t right to blame China! \\ \addlinespace 
& \textbf{Describe content as hateful or prejudicial} & you shouldn’t say that, it’s derogatory \\ \addlinespace 
& \textbf{Express solidarity with target entities} & Stand with Chinatown against racists. \\ 
\bottomrule
\end{tabularx} 
\caption{Taxonomies of counterspeech proposed by various authors. Both \citet{mathew2019thou} and \citet{conan-2019} add new categories to the classes proposed by \citet{benesch2016}. All the reported examples come from the relative papers, except for the \textit{Humor and sarcasm} example, which is taken from \citet{fanton2021human} dataset.}
\label{tab:cn_taxonomies}
\end{table*}

As mentioned in Section \ref{sec:taxonomies}, the most widely employed counterspeech taxonomy is the one proposed by \citet{benesch2016}. However, \citet{qian-etal-2019-benchmark} make a different distinction, based on the observed strategies adopted by the crowdworkers in their study. These consist of \textit{Identifying Hate Keywords}, \textit{Categorize Hate Speech}, \textit{Positive Tone Followed by Transitions}, and \textit{Suggest Proper Actions}. In particular, the strategy of \textit{Identifying Hate Keywords} is based on exhorting users to stop using inappropriate terms. \textit{Categorize Hate Speech} involves the classification of the hate speech into a specific category. \textit{Positive Tone Followed by Transitions} relies on showing empathy first and then proceeding to condemn the hateful text. Finally, with \textit{Suggest Proper Actions} a proactive suggestion is made to the user. 

Alternatively, \citet{vidgen2020detecting} propose a taxonomy where counterspeech are distinguished according to whether it \textit{Rejects the premise of abuse}, \textit{Describes content as hateful or prejudicial}, or \textit{Expresses solidarity with target entities}. Examples of all the identified strategies are shown in Table \ref{tab:cn_taxonomies}.\looseness=-1 

\subsection{Datasets for counterspeech-related tasks} \label{app:related_datasets}

In Table \ref{tab:related_datasets}, we make a non-exhaustive list of available datasets for the tasks described in Section \ref{sec:cn_related_tasks}.

\begin{table*}[htbp]
\small
\centering
\begin{tabularx}{\textwidth}{lXrXXX}
\toprule
\textbf{Dataset}   & \textbf{Task}              & \textbf{Size}  & \textbf{Interact.}                                                & \textbf{Coll.} & \textbf{Source}       \\
\midrule

\citet{lee-etal-2022-elf22}  & Counter-trolling    & 6,686  & Pairs  & Crawl. and Crowd.  & Reddit \\
\citet{chakravarthi2020hopeedi} & Hope speech & 59,354 & Single c. &  Crawl. & YouTube \\
\citet{garcia2023hope} & Hope speech & 1,650 & Single c. &  Crawl. & Twitter \\
\citet{palakodety2019hope} & Hope speech & 921,235 & Single c. &  Crawl. & YouTube \\
\citet{kim-etal-2022-prosocialdialog} & Prosocial dialogue & 58,137 & Dialog. & Hybr. & Morality-related data \\
\citet{logacheva2022paradetox} & Detoxification & 12,000 & Pairs & Crowd. & Toxic sentences data
\\
\citet{ung2022saferdialogues} & Feedback on safety failures & 7,881 & Dialog. & Hybr. & \citet{xu2021bot} \\
\citet{stammbach2020fever} & Misinformation countering & 67,687 & Triplets 
& Hybr. & \citet{thorne2018fever} \\
\citet{kotonya2020explainable} & Misinformation countering & 11,832 & Pairs 
& Crawl. & Fact-checking websites \\
\citet{ma2023characterizing} & Misinformation countering & 690,047 & Pairs & Crawl. & Twitter \\
\citet{alhindi2018your} & Misinformation countering & 12,836 & Triplets 
& Crawl. & 
\href{https://www.politifact.com/}{PolitiFact} \\
\citet{russo2023benchmarking} & Misinformation countering &8,289 & Triplets 
& Crawl. & Fact-checking websites \\
\citet{russo2023countering} & Misinformation countering & 11,990 & Triplets 
& Hybr. &
\href{https://fullfact.org}{Full Fact} \\

\bottomrule
\end{tabularx} 

\caption{Available datasets on tasks related to counterspeech. The data collected by \citet{russo2023benchmarking} are not distributed, but they share the code to replicate the data collection.}
\label{tab:related_datasets}
\end{table*}

\subsection{Annotators training procedure} \label{app:annot_training}
Recognizing, post-editing and writing counterspeech requires expertise and practice. When annotators do not have any previous experience incounterspeech they can be trained to acquire proficiency in the task of interest \cite{chung2020italian, vidgen2020detecting, vidgen2021introducing, fanton2021human, he2021racism, furman2022parsimonious, bonaldi2022human, gupta-etal-2023-counterspeeches, bonaldi-etal-2023-weigh}. The most employed procedure for annotators' training includes the following steps: 
\begin{itemize}
    \item[a)] reading and discussing NGO guidelines and public documentation describing the activity of counterspeech writing;
    \item[b)] reading both examples of counterspeech writing and of the specific task of interest (e.g. post-editing) performed by experts;
    \item[c)] practice the task on a subsample of examples;
    \item[d)] discuss disagreements with an expert.
\end{itemize}

This procedure can last from two to four weeks. Table \ref{tab:annot_training} summarises the steps undertaken by the studies explicitly describing how they trained the annotators. Furthermore, it is important to preserve the \textit{well-being} of the annotators, given the risks involved in working with hateful content. In particular, taking simple precautions like those suggested by \citet{vidgen2019challenges} is enough to safeguard the annotators' mental health. These precautions include explaining the prosocial purpose of the research, limiting the annotation time to no more than three hours per day, taking regular breaks, and having several meetings to allow for possible problems or distress to emerge. 

\begin{table}[ht!]
    \centering
    \small
    \resizebox{\columnwidth}{!}{
    \begin{tabular}{lcccc}
    \toprule
        \textbf{Study} & \textbf{a} & \textbf{b} & \textbf{c} & \textbf{d} \\
        \midrule
        \citet{chung2020italian} & \ding{51} &  \ding{51} & - & -\\
         \citet{he2021racism} & - & - & \ding{51} & \ding{51} \\ 
         \citet{vidgen2021introducing} & - & - & - & \ding{51} \\
         \citet{fanton2021human} & \ding{51} & \ding{51} & \ding{51} & \ding{51}\\
        \citet{bonaldi2022human} & \ding{51} & \ding{51} & \ding{51} & \ding{51}\\
        \citet{gupta-etal-2023-counterspeeches}  & \ding{51} & - & - & - \\
        \citet{bonaldi-etal-2023-weigh} & - & \ding{51} & -  & \ding{51}\\
        \bottomrule
    \end{tabular}}
    \caption{The steps for annotators' training in the studies that explicitly mention them, as described in \S\ref{app:annot_training}.}
    \label{tab:annot_training}
\end{table}

\end{document}